\documentclass[11pt,letterpaper]{article}
\usepackage[]{acl2017}
\usepackage{times}
\usepackage{latexsym}
\usepackage{pbox}
\usepackage{pstricks,pst-node,pst-tree}
\usepackage{multirow}
\usepackage{graphicx}
\usepackage{algorithm}
\usepackage{algpseudocode}
\usepackage{amsmath} 
\usepackage{booktabs}
\usepackage{tikz}
\usetikzlibrary{positioning}
\usepackage{comment}
\usepackage{amssymb}
\usetikzlibrary{arrows}
\usepackage{soul}
\usepackage{color,soul}
\usepackage{multirow}
\usepackage[normalem]{ulem}
\usepackage{graphicx,enumitem}
\usepackage{lipsum}%
\usepackage{tkz-graph}
\usepackage{newunicodechar}

\usepackage{float}
\newcommand{\argmax}{\mathop{\mathrm{argmax}}}
\usepackage[skip=2pt]{caption}
\usetikzlibrary{arrows}
\usepackage{array}
\aclfinalcopy 
\def\aclpaperid{***} 
\DeclareCaptionFormat{myformat}{\fontsize{9}{10}\selectfont#1#2#3}
\title{Semi-Supervised Affective Meaning Lexicon Expansion Using Semantic and Distributed Word Representations   }

\author{Areej Alhothali \\
  David R. Cheriton School \\of Computer Science \\
  University of Waterloo \\
  Waterloo, Ontario, N2L3G1 \\
  {\tt aalhotha@cs.uwaterloo.ca} \\\And
  Jesse Hoey \\
   David R. Cheriton School \\of Computer Science \\
  University of Waterloo \\
  Waterloo, Ontario, N2L3G1 \\
  {\tt jhoey@cs.uwaterloo.ca} \\}

\date{}

{%
\centering\addtocounter{figure}{1}
\begin{enumerate}[%
itemsep=2pt,parsep=0em,
label={(\alph*)},
ref={\thefigure.(\alph*)}
]}%
{\end{enumerate}\addtocounter{figure}{-1}}
\begin{document}
\maketitle
\begin{abstract}
In this paper, we propose an extension to graph-based sentiment lexicon induction methods by incorporating distributed and semantic word representations in building the similarity graph to expand a three-dimensional sentiment lexicon. We also implemented and evaluated the label propagation using four different word representations and similarity metrics. Our comprehensive evaluation of the four approaches was performed on a single data set, demonstrating that all four methods can generate a significant number of new sentiment assignments with high accuracy. The highest correlations ($\tau=0.51$) and the lowest error (mean absolute error $< 1.1\%$), obtained by combining both the semantic and the distributional features, outperformed the distributional-based and semantic-based label-propagation models and approached a supervised algorithm.

\end{abstract}

\section{Introduction}
\label{sec:Intro}
Sentiment analysis (SA) is a rapidly growing area of interest in natural language processing (NLP). Sentiment analysis is useful for a variety of important applications, such as recommendation system, virtual assistants, and health informatics.  Much SA relies on lexicons mapping words to sentiment, which are either manually annotated or automatically generated from a small set of seed words. Many researchers and companies have explored methods of expanding and re-generating sentiment lexicons to reduce the cost of manual annotation and to compensate for the lack of existing annotated data and the dynamic and fluctuating nature of human emotion. However, most sentiment lexicon expansion methods attach a polarity value (i.e., negative, positive, or neutral)~\cite{stone1968general} or real-valued one-dimensional scores~\cite{baccianella2010sentiwordnet} to the words. It is well known; however, that one dimension is insufficient to adequately characterise the complexity of emotion~\cite{Fontaine2007}.

In a large set of cross-cultural studies in the 1950s, Osgood showed that concepts carried a culturally dependent, shared affective meaning that could be characterised to a great extent using three simple dimensions of evaluation (good versus bad), potency (powerful versus powerless), and activity (lively versus quiet)~\cite{osgood1957measurement}. This {\em semantic differential} scale of evaluation, potency, and activity (EPA) is thought to represent universal and cross-cultural dimensions of affective meaning for words.

Based on this work, several three-dimensional sentiment lexicons have been manually labeled using surveys in different countries 
\cite{heise2010surveying}. Words in these lexicons are measured on a scale from $-4.3$ (infinitely bad, powerless, or inactive) to $+4.3$ (infinitely good, powerful, or lively)~\cite{berger2002new,heise2007expressive}.\footnote{The range $[-4.3,4.3]$ is a historical convention} In these surveys, participants are asked to rate identities (e.g., teacher, mother), behaviors (e.g., help, coach), adjectives (e.g., big, stubborn), institutions (e.g., hospital, school) or scenarios (e.g. combinations of identities, behaviours, adjectives and institutions)~\cite{heise2010surveying} on 5-item scales ranging from "Infinitely negative (e.g., bad/powerless/inactive)" to "Infinitely positive (e.g., good/powerful/active)", which are then mapped to the $[-4.3,4.3]$ scale. These manual annotation methods are labor-intensive, time-consuming and they produce a relatively small number of words. 

In this paper, we utilize the semantic and distributed words representation to expand these three-dimensional sentiment lexicons in a semi-supervised fashion. We also evaluated four different approaches of computing the affinity matrix using a semantic (dictionary-based) features, singular value decomposition word embedding, neural word embedding word vector, and combining both neural word embedding and semantic features. The highest results were obtained using the semantic and neural word embedding model with a rank correlation score $\tau=0.51$ on recreating two sentiment lexicon~\cite{warriner2013norms} and the General Inquirer~\cite{stone1968general}. The results also show that the highest rank correlation scores of the three dimension were for evaluation (E) while the lowest were for the potency (P). We also evaluated our induced EPA scores against some of the state-of-the-art methods in lexicon expansion, and our method shows an improvement in the $\tau$ correlation and F1 score over these algorithms. 

Our contributions are fivefold: 1) this is the first work that extensively examines methods of multidimensional lexicon expansion (we compute the evaluation, potency, and activity (valence, dominance, and arousal) scores instead of only computing the evaluative factor ( valence), 2) we propose a label propagation algorithm that is built upon both the semantic and distributed word representations, 3) we performed a comprehensive evaluation of four algorithms against a manually annotated dataset as well as a supervised learning algorithm, 4) we sample seed words from the corpus or dictionary instead of using the commonly used fixed seed words (e.g., good, bad, happy, sad etc.), 5) we created a significantly large three-dimensional lexicon of $\sim\!\! 3$M words that could be leveraged by researchers in fields of sentiment analysis and social science.   

Our proposed approaches 1) reduce the cost of manual annotation of sentiment lexicons; 2) integrate the affective meaning of today's' growing vocabulary (e.g., selfie, sexting), and 3) identify and incorporate the variance in attitudes towards words (e.g., same-sex marriage, abortion).

\section{Related Work}
\label{sec:Relatedwork}
The lexicon augmentation methods in this study were performed using variations of word representations and similarity metrics. This section provides a short background about the various vector space models that are used.
\subsection{Statistical language modeling}
Statistical language model (or vector space model (VSM)) is a distributional estimation of various language phenomena estimated by employing statistical techniques on real world data.
Representing language phenomena in terms of parameters has proven to be useful in various natural language processing (NLP), speech recognition, and information retrieval (IR) tasks. To capture the semantic or syntactic properties and represent words as proximity in n-dimensional space, several VSMs have been proposed ranging from the simple \emph{one-hot} representation that regards words as atomic symbols of the co-occurrence with other words in a vocabulary to a \emph{neural word embedding} that represents words in a dense and more compact representation.

The most commonly used word representation is the \emph{distributional word embeddings} representing word based on the co-occurrence statistics with other words in a document or corpora~\cite{harris1981distributional,firth1957synopsis}. The dimensionality of this sparse representation can be reduced using Singular value decomposition~\cite{eckart1936approximation}, Latent Semantic Analysis~\cite{landauer1997solution} or Principal Component Analysis~\cite{jolliffe2002principal}. 

\emph{Neural word embeddings} has recently gained a lot of attention in NLP and deep learning. 
\emph{Neural word embeddings} represent words in a low-dimensional, continuous space where each dimension corresponds to semantic or syntactic latent features.  Similar to distributional word embeddings,  \emph{neural word embeddings} are usually based upon co-occurrence statistics, but they are more compact, less sensitive to data sparsity, and able to represent an exponential number of word clusters~\cite{bengio2006neural}~\cite{mikolov2010recurrent,mikolov2011extensions}.

\subsection{Acquisition of Sentiment Lexicon}

Similar to other NLP tasks, sentiment lexicon induction methods can be achieved using two main approaches \emph{corpus-based} or \emph{thesaurus-based}. Turney and Littman~\cite{turney2003combining} proposed a \emph{corpus-based} lexicon learning method by first applying (TF-IDF) weighting on matrices of words and context, using SVD, and then computing the semantic orientation with a set of seed words. 

\emph{Thesaurus-based} methods use the lexical relationship such as the depth of a concept in taxonomy tree~\cite{wu1994verbs} or edge counting~\cite{collins1969retrieval} to build sentiment lexicons. Similar to Turney's PMI approach ~\cite{kamps2004using} they use WordNet based relatedness metric between words and given seed words. 

\emph{Semi-supervised graph-based models} that propagate information over lexical graphs have also been explored. The polarity-propagation or sense propagation algorithm induces sentiment polarity of unlabeled words given seed words (positive, negative) and the lexical relationships between them (e.g., word-net synonym, antonym)~\cite{Strapparava2004,esuli2006sentiwordnet}. Some researchers have developed a weighted label propagation algorithm that propagates a continuous sentiment score from seed words to lexically related words~\cite{Godbole2007}. Velikovich et al.~\cite {velikovich2010viability} proposed web-based graph propagation to elicit polarity lexicons. The graph is built upon a co-occurrence frequency matrix and cosine similarity (edges) between words and seed words (nodes). Then, both a positive and a negative polarity magnitude will be computed for each node in the graph which is equal to the sum over the max weighted path from every seed word (either positive or negative). 

Several recent studies have utilized \emph{word embeddings} to generate sentiment lexicons, such as a regression model that uses structured skip-gram 600 word embedding to create a Twitter-based sentiment lexicon~\cite{astudillo2015inesc}. Another study transforms dense word embedding vectors into a lower dimensional (ultra-dense) representation by training a two objective function gradient descent algorithms on lexicon resources~\cite{Rothe2016}. A recent study has also proposed a label propagation based model that uses word embedding, built using singular value decomposition (SVD) and PMI, to induce a domain-specific sentiment lexicon~\cite{hamilton2016inducing}. 

Few studies have looked at multidimensional sentiment lexicon expansion. 
Kamps et al.~\cite{kamps2004using} use a WordNet-based metric to elicit semantic orientation of adjectives. 
The generated lexicon was evaluated against the manually constructed list of Harvard IV-4 General Inquirer~\cite{stone1968general}. Kamps et al.'s work focuses only on adjectives and assigns them a binary value (either good or bad, potent or impotent, etc.). A  three-dimensional sentiment lexicon was extended using a thesaurus-based label propagation algorithm based upon WordNet similarity~\cite{alhothaligood}, and their results were compared against the Ontario dataset~\cite{mackinnon2006mean}. 

\section{Method}
\label{sec:method}

\normalfont
\subsection{Graph-based Label-Propagation }
Expanding sentiment lexicons using graph-based propagation algorithms was pursued previously and found to give higher accuracy in comparison with other standard methods~\cite{hu2004mining,andreevskaia2006semantic,rao2009semi}. To evaluate the effectiveness of graph-based approaches in expanding multidimensional sentiment lexicons, in this paper, we use the label propagation algorithm~\cite{Zhu2002,zhou2004learning},  combined with four methods for computing words vectors and word similarities.
The label propagation algorithms rely on the idea of building a similarity graph with labeled (seed words/paradigm words) and unlabeled nodes (words). The labels or scores of the known nodes (words) are then propagated through the graph to the unlabeled nodes by repeatedly multiplying the weight matrix (affinity matrix) against the labels or scores vector. 

Following the same principle, the graph label propagation algorithm in this paper: 1) creates a set of labeled  $L=(X_l,Y_l)$ and unlabeled data points or words $U=(X_u,Y_u)$ where $|U|+|L|=|V|$, V is all the words in the vocabulary set, $X$ is the word, and $Y$ is the sentiment (E, P, A scores) attached to that word; 2) constructs an undirected weighted graph $G=\{E, V, W\}$ where $V$ is a set of vertices (words), $E$ edges, $W$ is an $|V| \times |V|$ weight matrix ( where $w_{ij}\geq 0)$; 3) Compute the random walk normalized Laplacian matrix $\Delta=D^{-1}W$ (where $D$ is the degree matrix); 4) initializes the labeled nodes/words $Y_l$ with their EPA values, and the unlabeled nodes/words $Y_u$ with zeroes; 4) propagates the sentiment scores to adjacent nodes by computing $Y \gets \Delta Y$ (weighted by a factor $\alpha$) and clamps the labeled nodes $Y_l$ to their initial values $L$ after each iteration.

We implemented the label propagation algorithm using four different methods of computing affinity matrix and word representations. First,  a semantic lexicon-based approach in which the graph is built based upon the semantic relationship between words (Semantic lexicon-based Label propagation or SLLP). Second, a distributional based approach in which vocabulary and weights come from co-occurrence statistics in a corpus (corpus-based label propagation or CLP). Third, a neural word embeddings method (neural word embedding label propagation or NWELP), and fourth, a combination of semantic and distributional methods (semantic neural word embedding label propagation or SNWELP).  The following subsections describe these four different methods of label propagation.

\subsubsection{Semantic Lexicon-based Label Propagation (SLLP)}
\label{ssec:Lexicon induction}

The SLLP algorithm follows the general principle of the graph-based label propagation approach as described in the previous section, but the affinity matrix $W$ is computed using the semantic features obtained from semantic lexicons. Two semantic lexicons were used in this algorithm: WordNet dictionary (WN)~\cite{miller1995wordnet} and the paraphrase database (PPDB)~\cite{ganitkevitch2013ppdb}. The SLLP algorithm constructs the vocabulary $V$ from the words of the dictionaries and computes and normalizes the weight matrix $W$ using the synonyms relationship between words. The semantic-based similarity $w_{i,j}$ of any pair of words $x_i$ and $x_j$ in the vocabulary $V$ is calculated as follows:
\begin{equation}
    w_{i,j}=
    \begin{cases}
      1.0 & \text{if}\ x_j  \text{ is a synonym of } x_i \\
      0.0 & \text{otherwise}
    \end{cases}
\label{eq:semanic}
  \end{equation}

\subsubsection{Corpus-based Label Propagation (CLP)}

Corpus-based label propagation (CLP) is one of the most commonly used methods for sentiment lexicon generation that uses the co-occurrence statistics aggregated from different corpora (news articles, Twitter, etc.) to build the similarity graph in the label propagation algorithms. We used an n-gram features from the signal media (SM) one million news articles dataset which contains $\sim\!\!265$K blog articles and $\sim\!\!734$K news articles ~\cite{Signal1M2016} and the North American News (NAN) text corpus~\cite{graff1995north} which has $\sim\!\!931$K articles from a variety of news sources.

The co-occurrence matrix $R$ was computed on a window size of four words. Bigrams with stop words, words less than three letters, proper nouns, non-alpha words, and the bigrams that do not occur more than ten times were filtered out. These heuristics reduce the set into $\sim\!\!80$k and $\sim\!\!40$k, for SM and NAN corpora, respectively.
We constructed the word vectors by computing the smoothed positive point-wise mutual information(SPPMI)~\cite{levy2015improving} of the co-occurrence matrix $R$. This smoothing technique reduces the PMI's bias towards rare words and found to improve the performance of NLP tasks~\cite{levy2015improving}. 

\begin{equation}
  \bf {SPPMI}_{ij}=max \{(log_2 \frac {p(\bf{w_i},\bf{w_j})}{p(\bf {w_i})p_\alpha(\bf{w_j})},0)\}
\end{equation}

where $ p(\bf{w_i},\bf{w_j})$ is the empirical co-occurrence probability of a pair of words $\bf {w_i}$ and $\bf {w_j}$ and $p(\bf {w_i})$ and $p_\alpha(\bf{w_j})$ are the marginal probability of  $\bf{w_i}$ and the smoothed marginal probability of $\bf{w_j}$, respectively. We use $\alpha=0.75$ as it is found to give better results~\cite{levy2015improving}~\cite{mikolov2013distributed} and we also experiment with the unsmoothed PPMI.
 The SPPMI matrix is then factorized with truncated Singular
Value Decomposition (SVD)~\cite{eckart1936approximation} as follows:
\begin{equation}
\bf SPPMI=U *\Sigma *V^T
\end{equation}
We take the top $k$ rows of $U$ as the word representation or word vector (we used k=300):
 \begin{equation}
\bf W_{SVD}=U_k 
\end{equation}
The affinity matrix is then computed as: 
\begin{equation}
 \bf w_{ij} = \cos({\bf v_i},{\bf v_j})= {{\bf v_i} {\bf v_j} \over \|{\bf v_i}\| \|{\bf v_j}\|}  ,  \\\ \forall v_i, v_j \in W_{SVD}
\label{eq:cossim}
\end{equation}

\subsubsection{ Neural Word Embeddings Label-propagation (NWELP)}
\label{sec:WELP}

This method uses word embeddings (word-vectors) that capture syntactic and semantic properties.
We use two pre-trained word embedding models that are trained on co-occurrence statistics. We used skip-gram word vector (SG)~\cite{mikolov2013distributed} that is trained on a skip-gram model of co-occurrence statistics aggregated from Google News dataset and Global vector for word representation(GloVe)~\cite{pennington2014glove} which have been trained on co-occurrence statistics aggregated from  Wikipedia. 
The vocabulary $V$ in this algorithm is all words in the word embeddings set (we filtered out non-alpha words and words that contain digits), and the affinity matrix $W$ is computed using the cosine similarity (Equation~\ref{eq:cossim}) between word vectors (each $v_i \in V$ is a 300 dimensional vector).

\subsubsection{Semantic and Neural Word Embeddings Label-propagation (SNWELP)}
To improve the results of the NWELP algorithm, we propose the  SNWELP,  a model that combines both semantic and distributional information obtained from the neural word embedding models and a semantic lexicon (a dictionary). The SNWELP algorithm constructs the affinity matrix $W$ 
using the neural word embeddings features (SG or GloVe) and semantic features obtained from a semantic lexicon (WN or PPDB).  In this case, $V$ is intersection between the words in lexicon and the word in the filtered embeddings set,  $W$ is the averaged cosine similarity scores (Equation~\ref{eq:cossim}) of the neural and the semantic word representations (Equation~\ref{eq:semanic}). 
 
\subsection{Sampling Methods\label{sec:sampling}}
Choosing the labeled words (also called {\em paradigm} or {\em seed} words) in the graph-based label propagation methods is one of the critical factors. We used two methods: 
1) fixed seed sets (\textit{fixed-paradigms}), and 2) words sampled from the vocabularies $V$ used in the label propagation algorithm (\textit{vocabulary-paradigms}). The \textit{fixed-paradigms} set was chosen from Osgood et al's~\cite{osgood1957measurement} research as shown in Table~\ref{seedwords-table} while the \textit{vocabulary-paradigms} set was randomly sampled from the corpus' vocabulary for words with the highest and lowest EPA values (words with E, P or A $ \leq -2.5$ or $\geq 2.5$). The objective is to use words at extremes of each dimension E,P, and A, as paradigm words in order to propagate these highly influencing EPA throughout the graph. The seed words contribute to no more than $~1\%$ of all words in each algorithm. We tested with the \textit{ fixed-paradigms} sets, but the results of the \textit{vocabulary-paradigms} were significantly better.  

\begin{table}[t]

\resizebox{\linewidth}{!}{%

\begin{tabular}{l}
\hline \bf EPA Seed words\\
\hline
  E+=\{good, nice, excellent, positive, warm, correct, superior\}\\
 E-=\{bad, awful, nasty, negative, cold, wrong, inferior\}\\ 
P+=\{powerful, strong, potent, dominant, big, forceful ,hard\}\\ 
  P-=\{powerless, weak, impotent, small, incapable, hopeless, soft\}\\ 
 A+=\{active, fast, noisy, lively, energetic, dynamic, quick, vital\}\\ 
 A-=\{quiet, clam, inactive, slow, stagnant, inoperative, passive\}\\ \hline

\end{tabular}
}
\caption{\label{seedwords-table} Osgood's  fixed seed words (+ positive word and - negative words) }

\end{table}

\subsection{Evaluation Metrics}
\label{evaluation}
To evaluate the effectiveness of the algorithm in generating a multidimensional sentiment lexicon, we chose the most recent manually-annotated affective dictionary~\cite{warriner2013norms} as baseline. We use the ~\cite{warriner2013norms} dictionary in the lexicon induction procedure by sampling the paradigm words from it and we compare the generated lexicon against it. We randomly divided the~\cite{warriner2013norms} affective dictionary (\textit{original-EPA}) into \textit{EPA-training} (third of the set equal to $5566$ words) and \textit{EPA-testing} (two-thirds of the set equal to $8349$ words). The seed words for all algorithms are sampled from the \textit{EPA-training} set only, and all results are presented on the \textit{EPA-testing} set. 

The EPA scores of ~\cite{warriner2013norms} initially range $\in[1 ,9]$ and we rescaled them to $ \in[- 4.3 ,+4.3]$ to follow the same EPA scale used in the other lexicons we have considered~\cite{heise2010surveying}. The $[- 4.3 ,+4.3]$ scale is the standard scale used by most of the researchers in the sociology field who study or measure individuals' emotions towards terms. 

Four evaluation metrics were used to compare the induced EPA (\textit{EPA-induced}) against the manually annotated EPA (\textit{EPA-testing}): mean absolute error (MAE), Kendall $\tau$ rank correlation, F1-binary (positive and negative), and F1-ternary (positive, neutral, and negative).  We used F1-binary to evaluate the binary classification performance of the model (positive $\geq0$ and negative $<0$ ) and similar to most recently proposed studies in the field~\cite{hamilton2016inducing}, we computed F1-ternary to measure the ternary classification accuracy: positive $\in (1,4.3]$,  neutral $\in[-1,1]$, and negative $\in [-4.3,-1)$. 
To calculate the F1-ternary, we used the \emph{class-mass normalization} (CMN) methods~\cite{zhu2003semi} that rescale the predicted label ($\hat{y}_{i,l}$) for a point $x_i$ by incorporating the class prior as follows :
$$\argmax_l \ \ w_l \ \hat{y}_{i,l} $$
 where $w_l$ is the label mass normalization which is equal to ${p_l}/{m_l}$ where $p_l$ is the prior probability of a label $l$ (computed from the labeled data), and $m_l$ is the estimated weight of label $l$ over the unlabeled sets. 
This scaling method is known to improve the results in comparison with the typical decision function $\argmax_l \ \ \hat{y}_{i,l}.$

\subsection{Baseline and State-of-the-art Comparison}
We compared our induced results against some of the standard state-of-art algorithms for inducing the valence (evaluation scores). We implemented the PMI-IR algorithm proposed by~\cite{turney2003combining} which estimates the sentiment orientation (either positive or negative) of a word by computing the difference between the strength of the word associations with positive paradigm words and with negative paradigm words using the co-occurrence statistics aggregated from search engines' results. We also compare our results against the reported results of~\cite{Rothe2016}'s orthogonal transformation of word vectors,
 and a label spreading algorithm trained on ( a domain-specific) SVD word vector model~\cite{hamilton2016inducing}. We also experimented with the retrofitted word vector model that improves the neural word embedding vectors using semantic features obtained from the lexical resources (WN, PPDB)~\cite{faruqui2014retrofitting}. 

To make a fair comparison, we implemented our label propagation algorithm and the retrofitted word vector approach~\cite{faruqui2014retrofitting} to recreate the General Inquirer lexicon~\cite{stone1966general} with valence score $\in R$ from~\cite{warriner2013norms} lexicon to compare our results to~\cite{hamilton2016inducing} and~\cite{Rothe2016}.We also ignored the neutral class and used the same seed set used by~\cite{hamilton2016inducing} and other researchers in the field. 
We also compare all the results against the EPA scores obtained from a supervised learning algorithm. We trained a support vector regression (SVR) model on a co-occurrence statistics model derived from the skip-gram word embedding model (SG)~\cite{mikolov2013distributed} and sentiment lexicon resource~\cite{warriner2013norms}. The SVR model uses RBF kernel with  $C=10$, and $\gamma=0.0$ for training and is trained on the full training set (EPA-training). 


\begin{table*}[t]
\setlength{\tabcolsep}{3pt}
\begin{center}
\begin{tabular}{|c|c|c|ccc|ccc|ccc|ccc|}
\hline  \footnotesize \bf Method& \bf \footnotesize   Corpus &  \footnotesize \bf W & \multicolumn{3}{c}{ \footnotesize \bf $\tau$ }  & \multicolumn{3}{c} { \footnotesize \bf F1-binary}& \multicolumn{3}{c}{ \footnotesize \bf F1-ternary } & \multicolumn{3}{c|}{ \footnotesize \bf MAE}  \\
 \cline{4-15}

  & & & \footnotesize E& \footnotesize P& \footnotesize A  &  \footnotesize E& \footnotesize P& \footnotesize A  & \footnotesize E& \footnotesize P& \footnotesize A  & \footnotesize E& \footnotesize P& \footnotesize A \\\hline

 \multirow{2}{*}{ \footnotesize \bf  CLP} &\footnotesize   \bf SM& \footnotesize  5,109& \footnotesize   0.219 & \footnotesize 0.0263  & \footnotesize 0.162  & \footnotesize  0.53 &  \footnotesize 0.44 & \footnotesize   0.56     & \footnotesize 0.42 &\footnotesize 0.45   &\footnotesize 0.44 & \footnotesize   1.1&\footnotesize    1.09 &\footnotesize  0.85  \\

  & \footnotesize \bf  NAN& \footnotesize  4,653 & \footnotesize   0.122 & \footnotesize  0.060 & \footnotesize 0.084   & \footnotesize0.51 & \footnotesize  0.54 & \footnotesize 0.54  & \footnotesize 0.50  &\footnotesize  0.42 &\footnotesize 0.45   & \footnotesize 1.3  &\footnotesize   1.0 &\footnotesize  0.99  \\  
\hline

\multirow{2}{*}{ \footnotesize  \bf SLLP} &\footnotesize \bf WN & \footnotesize 4,801& \footnotesize   0.388& \footnotesize  0.244& \footnotesize 0.329 & \footnotesize 0.72 & \footnotesize  \bf 0.83 & \footnotesize   0.73& \footnotesize 0.65&\footnotesize 0.60  &\footnotesize 0.75 & \footnotesize 0.91 &\footnotesize \bf 0.79 &\footnotesize  \bf 0.71  \\ 
&\footnotesize \bf PPDB & \footnotesize 4,621 & \footnotesize   0.391& \footnotesize  0.181& \footnotesize 0.309 & \footnotesize 0.73 & \footnotesize  0.76 & \footnotesize   0.71& \footnotesize 0.62&\footnotesize 0.60  &\footnotesize  0.65& \footnotesize 0.92 &\footnotesize   0.89  &\footnotesize  0.79  \\
\hline
\

  \multirow{2}{*}{ \footnotesize \bf NWELP} & \footnotesize \bf  SG & \footnotesize 8,072  & \footnotesize  0.437 & \footnotesize 0.283  & \footnotesize  0.350 & \footnotesize   0.70& \footnotesize    0.80 & \footnotesize    0.67 & \footnotesize \bf  0.69 &\footnotesize \bf 0.65   &\footnotesize   \bf 0.79*& \footnotesize \bf 0.84 &\footnotesize 1.08 &\footnotesize   0.88 \\
&\footnotesize \bf GloVe& \footnotesize  4,867& \footnotesize  0.430& \footnotesize 0.113 & \footnotesize 0.357 & \footnotesize  0.73 & \footnotesize   0.81& \footnotesize 0.70&\footnotesize  0.68 & \footnotesize 0.64&\footnotesize   0.78&\footnotesize  1.09&\footnotesize  1.07&\footnotesize  0.84\\
\hline 
  \multirow{4}{*}{ \footnotesize \bf SNWELP}
 &\footnotesize \bf  PPDB+GloVe  & \footnotesize  4,867& \footnotesize  0.434  & \footnotesize 0.209&  \footnotesize 0.360 & \footnotesize  0.74 & \footnotesize 0.81 & \footnotesize 0.70& \footnotesize 0.68 &\footnotesize 0.64   &\footnotesize 0.77& \footnotesize  1.09 &\footnotesize 1.07 &\footnotesize 0.84 \\

 &\footnotesize \bf  \footnotesize WN+GloVe  & \footnotesize  4,867 & \footnotesize 0.445  & \footnotesize 0.220& \footnotesize 0.366 & \footnotesize 0.75  & \footnotesize 0.82& \footnotesize 0.71 & \footnotesize 0.68&\footnotesize 0.64   &\footnotesize 0.78& \footnotesize 1.07    &\footnotesize 1.05& \footnotesize 0.84   \\
&\footnotesize \bf  PPDB+SG  & \footnotesize 4,818  & \footnotesize \bf  0.510  & \footnotesize 0.284& \footnotesize 0.459 & \footnotesize  \bf 0.76 & \footnotesize 0.80& \footnotesize  \bf 0.75& \footnotesize 0.68&\footnotesize 0.64   &\footnotesize 0.78 & \footnotesize  1.10 &\footnotesize 0.97 &\footnotesize 0.84 \\
  &\footnotesize \bf  WN+SG  & \footnotesize 5,367  & \footnotesize \bf 0.510  & \footnotesize  \bf 0.291 & \footnotesize \bf 0.461& \footnotesize \bf 0.76 & \footnotesize 0.80& \footnotesize \bf 0.75& \footnotesize 0.68 &\footnotesize 0.64   &\footnotesize 0.78 & \footnotesize  1.10 &\footnotesize  0.95&\footnotesize 0.83  \\

\hline

 \multirow{1}{*}{\footnotesize \bf SL}& \footnotesize \bf  WESVR & \footnotesize 8,271  & \footnotesize   \bf 0.628* & \footnotesize  \bf0.422* & \footnotesize   \bf0.500* & \footnotesize   \bf0.83*& \footnotesize   \bf0.84*& \footnotesize   \bf0.78*& \footnotesize  \bf 0.72* &\footnotesize  \bf 0.65* &\footnotesize  0.68  & \footnotesize   \bf 0.60* &\footnotesize   \bf 0.60*&\footnotesize   \bf 0.56* \\

\hline
\end{tabular}
\end{center}
\caption{\label{LP-results2}  The results of the label propagation algorithms and the supervised learning (SL) method (support vector regression (WESVR)) using the \textbf{sampled seed words} in comparison with the ground truth EPA values (Method= the algorithm used for lexicon induction, W= the number of the induced words that has label in the dictionary,  $\tau$ = Kendall's $\tau$ correlation, F1-binary= F1 measure of the binary classification, F1-ternary= F1 scores of the ternary classification, MAE=Mean Absolute Error). The highest scores of the label propagation algorithms are in a \textbf{boldface}. The highest scores of all the algorithm are in \textbf{boldface*}. }
\end{table*}

\section{Results}
\label{sec:results}
In this section, we present the results of comparing the induced EPA scores using the label propagation algorithms against their corresponding values in the \textit{EPA-testing}. 
As shown in Table~\ref{LP-results2},  using SVD word embeddings in the CLP algorithm generated the lowest ranking correlation $\tau$ and the highest error rate (MAE) in comparison with the other label propagation methods. The results of comparing the induced EPA scores against their true values in the testing set show that the MAE ranged between 0.99 and 1.3 and the ranking correlation \footnote{The p-value for all the reported $\tau$ scores are less than 0.001} $\tau$ was less than 0.2 using cosine similarity and hard clamping ( $\alpha =1.0$) assumption. 
We also experimented with the unsmoothed point-wise mutual information (PPMI), but there was not a significant difference between the smooth and the unsmoothed PMI. We also tried different dimension of the SVD word vector k=100 and k=300, but there was no significant difference between them as well.

The results of the SLLP algorithm that uses the semantic features obtained from either WN or PPDB  lexicons generated a total of $\sim\!\! 50$K words, where $\sim\!\! 4$K words are in the testing set (\textit{EPA-testing}). The results of comparing the induced EPA scores to their corresponding values in the testing set (\textit{EPA-testing}) show that the MAE  less than $1.0$, F1-binary  greater than $0.70$, F1-ternary greater than or equal to $0.60$, and the ranking correlation  $\tau \geq 0.2 $ suggesting that there is a reasonable degree of agreement between the induced EPA score using dictionary-based features and the manually labeled EPA values. 

The $\tau$ correlations scores show that neural word embedding label propagating NWELP outperformed the semantic based, and corpus-based label propagation algorithms, as shown in Table~\ref{LP-results2}. The MAE and F1-scores of the semantic-based and neural word embedding label propagation were close. The MAE of the neural word embedding ranged from 0.84 to 1.09, F-1 scores were between 0.67 and 0.80, and $\tau$ ranged from $0.1$ to $0.44$. Comparing the results of the two pre-trained neural word embedding shows that the skip-gram based (SG) model performed better than (GloVe).  We experimented with different thresholds ($0.0$, $0.3$, and $0.5$) of the cosine similarities and the result using different threshold varied a lot in respect to the number of induced words and the accuracy. Higher thresholds provided more accurate results and less noise in the results, but with less number of induced words. The reported results in Table~\ref{LP-results2} and~\ref{all-results} are using cosine similarity threshold equal to $0.0$ since the adjacency matrix of both SG and Glove contain negative values. Combining the semantic and neural word embedding features improved the results with $\tau$ ranged between $ 0.43$ and $0.51$ and MAE $\leq 1.1$ for the evaluation scores (E). The results of the supervised  SVR model significantly outperformed the results obtained from the semi-supervised method with $\tau$ equal to 0.628, 0.422 and 0.500 for E, P, and A, respectively, F-1 scores equal to 0.83, 0.84, and 0.78, and MAE close to 0.6, but the results of the SNWELP were comparable. 

Comparing the results across the different affective dimensions (E,P, and A) shows that the rank correlation $\tau$ of comparing the potency (P) to their counterpart scores in testing set was low in comparison with the scores of evaluation (E) and activity (A) in both the semi-supervised algorithms and the supervised algorithm. While the rank correlation $\tau$ of the evaluation (E) scores were the highest in all the algorithms which indicate that words with similar word embeddings have a similar evaluation score. Table~\ref{CLP-words} shows some of the induced EPA scores and their corresponding values in~\cite{warriner2013norms} dataset. The table also shows some examples of the same words and their induced EPA scores using different word representations. Comparing our induced evaluation scores (E) with some of the state-of-the-art methods, as shown in Table~\ref{all-results}, indicates that our label propagation algorithms significantly performed better than~\cite{turney2002}'s unsupervised method. The result also shows that semantic neural word embedding (SNWELP) model outperformed~\cite{Rothe2016} and~\cite{hamilton2016inducing} approaches. Also, the neural word embedding and semantic neural word embedding algorithms perform better than the label propagation that uses the retrofitted word vector (the reported results are of the improved skip-gram model (SG) using semantic features obtained from wordnet (WN))~\cite{faruqui2014retrofitting}. 

\begin{table}[t]

\scalebox{0.9}{
\begin{tabular}{|c|c|c|c|c|}
\hline   \bf Method&  $ \bf{ \tau}$   &  \small  \bf F1-ternary&  \small  \bf ACC  \\  
\hline

   \small  \bf SNWELP (SG+WN)   &   0.51    &  0.67   & 0.94\\
   \small   \bf \citep{hamilton2016inducing} &   0.50 &  0.62 &   0.93\\
   \small  \bf NWELP (SG) &  0.48 &   0.67&   0.94 \\  
 \small  \bf \citep{Rothe2016} &    0.44&   0.59&  0.91 \\
\small \bf \citep{faruqui2014retrofitting}   & 0.40    & 0.62  &  0.84    \\ 
  \small  \bf \citep{turney2002}  &   0.14 &  0.47 &   0.55\\ 

\hline
\end{tabular}}

\caption{\label{all-results}  The results of comparing evaluation (E) of the \textbf{General Inquirer} induced lexicon using \textbf{ the pre-trained} Neural Word Embeddings label propagation (NWELP) and  Semantic Neural Word Embeddings label propagation (SNWELP) and \textbf {fixed seed words} with the results reported by the state of the are results in method in lexicon induction ( $\tau$ = Kendall's $\tau$ correlation, ACC= the binary accuracy, F1= the ternary F-measure)}
\end{table}

\begin{table}[t]
\setlength{\tabcolsep}{3pt}
\begin{center}
\scalebox{0.9}{
\begin{tabular}{|c|c|c|c|}

\hline \footnotesize  \bf Word &\footnotesize  \bf  Method &\footnotesize  \bf  Induced EPA  &\footnotesize  \bf  True EPA  \\ \hline
 \footnotesize injustice &\footnotesize WN &\footnotesize [-1.9,  0.3, -1.7]  & \footnotesize [-2.7, 1.6, -1.86] \\
  \footnotesize injustice&\footnotesize GloVe &\footnotesize [-1.3,  1.4 , -1.8] & \footnotesize [-2.7, 1.6, -1.86] \\
  \footnotesize injustice&\footnotesize GloVe+WN &\footnotesize [-1.4,  0.2, -1.3] & \footnotesize [-2.7, 1.6, -1.86] \\
\footnotesize injustice &\footnotesize SG+ WN &\footnotesize [-1.9,  0.3, -1.7 ]* & \footnotesize [-2.7, 1.6, -1.86] \\
\footnotesize evil   &\footnotesize PPDB&\footnotesize [-1.3  ,  0.05, -1.1] & \footnotesize [-2.9, 0.7, -1.5] \\
\footnotesize evil  &\footnotesize GLoVe&\footnotesize [-2.1,  2.5, -3.1] & \footnotesize [-2.9, 0.7, -1.5] \\
\footnotesize evil   &\footnotesize GLoVe+PPDB&\footnotesize [-1.7 , 0.08, -1.2] & \footnotesize [-2.9, 0.7, -1.5] \\
\footnotesize evil   &\footnotesize SG+PPDB&\footnotesize [-2.1,  0.1, -1.5] & \footnotesize [-2.9, 0.7, -1.5] \\
\footnotesize successful&\footnotesize SG&\footnotesize[ 2.15,  0.04,  1.6]& \footnotesize[2.97, 0.09, 2.9]\\
\footnotesize successful&\footnotesize SG+PPDB&\footnotesize[ 2.5, -0.6,  2.0]& \footnotesize[2.97, 0.09, 2.9]\\

\hline

\end{tabular}}
\end{center}
\caption{\label{CLP-words} Some example of the induced EPA and their EPA ratings from Original-EPA-lexicon and the induced EPA values using label propagation and different word representations WN=wordnet, parahprese-database=PPDB, SG =skip-gram word vector, and GLoVe= the global vector for word representation. The starred example * show no change after adding the neural word vector features.}
\end{table}

\section{Discussion}
\label{sec:diss}
Sentiment analysis is a feature engineering problem in which sentiment lexicons play a significant role in improving the model accuracy. One of the challenges of sentiment analysis is the increasing number of new words and terms in the social media or news resources (e.g., selfie, sexting, photobomb,etc.) 
  that do not have a sentiment score attached to them. Also, there is a need to measure the variance in human attitudes towards some terms over a period of time (e.g., homosexuality, abortion) and to explore other dimensions of humans' emotions.  
To overcome these limitations, reduce the cost of manual annotation, and increase the number of the annotated terms, we propose an extension and an evaluation of corpus and thesaurus-based algorithms to automatically induce a three-dimensional sentiment lexicon.

 Similar to any NLP applications, the vast majority of the work in lexicon induction uses distributional word representations (corpus-based statistics). In this study, the corpus-based label propagation (CLP) algorithm generated the least accurate results. Also, despite the viability of distributional word representations, exactly what syntactic and semantic information it captures is hard to determine, and not clear whether it is relevant for sentiment at all. 
 
The semantic lexicon-based label propagation (SLLP) was better than CLP.  However, there are also some limitations of using the dictionary based approach 1) the synonym relationship can only be computed between words of the same part of speech, 2) the dictionary has a limited number of words and does not include words that are used in the social media and internet in general.
 
Only one study have experimented with neural word embedding label propagation to expand the one-dimensional sentiment lexicon~\cite{hamilton2016inducing} with only reporting the result of using SVD word embedding model. In our study, we report the results of using different neural word embedding models. The results show that our neural word embedding model performed better than the SVD word vector approach. These findings require further analysis and assessment on different corpora. 

The results of combining both the semantic and neural word embedding (NWELP) was better than the corpus-based or semantic lexicon-based algorithms. The semantic neural word embedding provided a higher rank correlation scores and a slighting lower MAE in comparison with the semantic lexicon and neural word embedding-based algorithms. The results of the semantic neural label propagation algorithm are also comparable with those generated using a supervised learning algorithm (SVR) trained on word embeddings and a sentiment lexicon. Using the semi-supervised algorithm; however, does not require a large training dataset and allows to annotate the words independently from the previously human-coded lexica.

\section{Conclusion}
\label{sec:con}

In this study, we propose an extension to the graph-based lexicon induction algorithms to expand sentiment lexicons and explore other dimensions of sentiments. This study to the best of our knowledge is the first work that expands a multi-dimension sentiment lexicon and the first to incorporates both the semantic and neural word representations in the label propagation algorithm. We also provided an extensive evaluation of label propagation algorithms using a variety of word representations that have been found to provide higher accuracy in many NLP tasks in comparison with other standard methods. The results show that the word semantic neural word embedding label propagation generates the highest correlations 
  compared with the corpus-based, semantic lexicon-based, and neural word embedding algorithms.


\bibliography{NAA2015}
\bibliographystyle{acl_natbib}


\end{document}